\documentclass[11pt]{article}

\usepackage[final]{acl}
\usepackage{times}
\usepackage{latexsym}

\usepackage[T1]{fontenc}

\usepackage[utf8]{inputenc}

\usepackage{microtype}

\usepackage{inconsolata}

\usepackage{graphicx}

\usepackage{microtype}
\usepackage{hyperref}
\usepackage{url}
\usepackage{booktabs}
\usepackage{multirow}
\usepackage{array}
\usepackage{subcaption}
\usepackage{longtable}
\usepackage{marvosym}
\usepackage{amsmath,amsfonts,bm}
\usepackage{subcaption}
\usepackage{stfloats}
\usepackage{placeins}

\usepackage{tabularx}
\usepackage{caption}

\title{Retrieval Improvements Do Not Guarantee Better Answers: A Study of RAG for AI Policy QA}

\author{
 \textbf{Saahil Mathur\textsuperscript{1}},
 \textbf{Ryan David Rittner\textsuperscript{1}},
 \textbf{Vedant Ajit Thakur\textsuperscript{2}},
 \textbf{Daniel Stuart Schiff\textsuperscript{2}},
 \textbf{Tunazzina Islam\textsuperscript{1}}\\
\\
 \textsuperscript{1}Department of Computer Science, Purdue University, West Lafayette, IN 47907 \\
 \textsuperscript{2}Department of Political Science, Purdue University, West Lafayette, IN 47907
}

\begin{document}
\maketitle
\begin{abstract}

Retrieval-augmented generation (RAG) systems are increasingly used to analyze complex policy documents, but achieving sufficient reliability for expert usage remains challenging in domains characterized by dense legal language and evolving, overlapping regulatory frameworks. We study the application of RAG to AI governance and policy analysis using the AI Governance and Regulatory Archive (AGORA) corpus, a curated collection of $947$ AI policy documents. Our system combines a ColBERT-based retriever fine-tuned with contrastive learning and a generator aligned to human preferences using Direct Preference Optimization (DPO). We construct synthetic queries and collect pairwise preferences to adapt the system to the policy domain. Through experiments evaluating retrieval quality, answer relevance, and faithfulness, we find that domain-specific fine-tuning improves retrieval metrics but does not consistently improve end-to-end question answering performance. In some cases, stronger retrieval counterintuitively leads to more confident hallucinations when relevant documents are absent from the corpus. These results highlight a key concern for those building policy-focused RAG systems: improvements to individual components do not necessarily translate to more reliable answers. Our findings provide practical insights for designing grounded question-answering systems over dynamic regulatory corpora.
\end{abstract}

\section{Introduction}
Artificial intelligence (AI) governance is evolving rapidly as governments and regulatory bodies introduce a diverse array of new laws, guidelines, and standards for AI systems \cite{wang2025artificial,maslej2025artificial,zhang2020us}. These policy documents are often lengthy, legally dense, and are distributed across numerous jurisdictions, making it difficult to analyze and compare across them. Resources such as the AI Governance and Regulatory Archive (AGORA) \cite{arnold2024introducing} dataset provide structured collections of AI policy documents, but extracting insights from these materials still requires substantial manual effort. Automated question-answering systems could help researchers and policymakers navigate this growing body of regulation.

Large language models (LLMs) \cite{brown2020language} provide powerful tools for analyzing complex text, but they often struggle with legal and regulatory documents due to domain-specific terminology, conceptual ambiguity, and nested references \cite{dahl2024large,henderson2022pile,askari2022expert}. Moreover, when applied directly to policy corpora, LLMs may generate fluent but unsupported claims \cite{tang2023policygpt}. Retrieval-augmented generation (RAG) \cite{lewis2020retrieval} addresses this limitation by grounding responses in retrieved documents \cite{pipitone2024legalbench}, yet the effectiveness of RAG depends heavily on both retrieval quality and generation alignment.

Despite recent advances in retriever training and preference-based alignment, it remains unclear whether improvements to individual RAG components consistently translate into better end-to-end question answering performance, particularly in complex and high-stakes domains. Domains like AI governance corpora present particularly serious challenges, given dense legal language, sometimes-ambiguous policy and technical jargon, as well as evolving and cross-referenced regulatory coverage across sectors and jurisdictions.

In this work, we investigate how domain adaptation affects RAG systems for AI policy question answering\footnote{Our code and data are publicly available here: \url{https://github.com/smathur23/agora} }
We construct a RAG pipeline over the AGORA corpus using a ColBERT-based retriever \cite{santhanam-etal-2022-colbertv2} and a generator aligned to human preferences via Direct Preference Optimization (DPO) \cite{rafailov2023direct}. The retriever is fine-tuned using contrastive learning with synthetically generated queries and manually labeled examples, while the generator is aligned using pairwise preference data collected from policy-focused question-answer tasks.

Our experiments evaluate retrieval performance, answer relevance, and response faithfulness. We find that while retriever fine-tuning improves retrieval metrics, it does not consistently improve end-to-end question answer performance. In some cases, stronger retrieval produces more confident hallucinations when relevant documents are absent from the corpus. These findings highlight an important challenge for policy-focused RAG systems: improvements to individual components do not necessarily translate into more reliable grounded responses. 
Our contributions are:
\newline
(1) An empirical study of retrieval-augmented generation for question answering over AI governance documents in the AGORA corpus.
\newline 
(2) A domain-adapted RAG pipeline combining contrastive retriever fine-tuning and preference-based generator alignment for policy analysis tasks.
\newline 
(3) An analysis showing that improvements in retrieval metrics may not translate into better end-to-end question answering performance, and can increase confident hallucinations when the underlying corpus lacks coverage.
\begin{figure}[t]
    \centering
    \includegraphics[width=1\columnwidth]{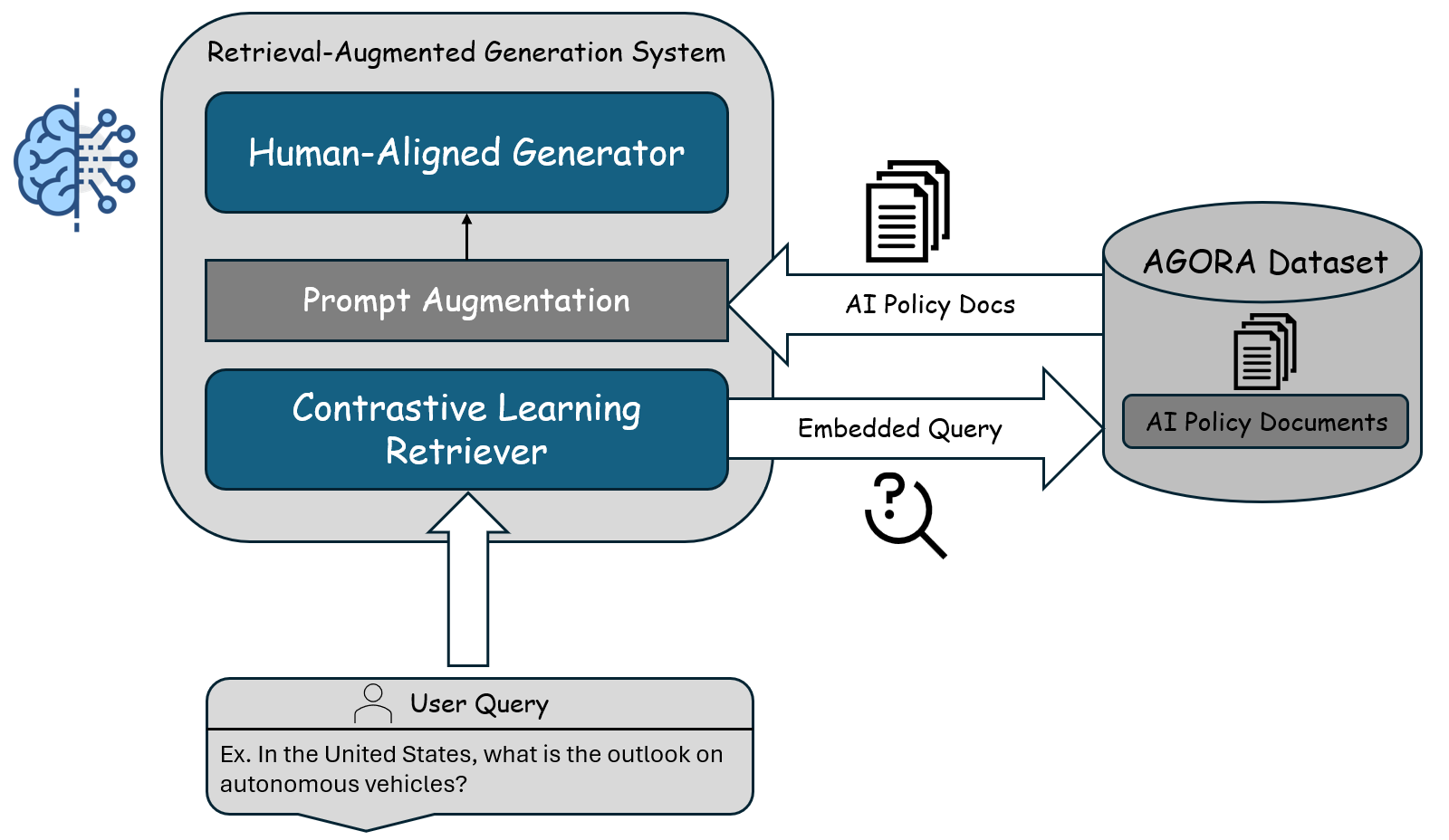}
    \caption{Overview of the AI policy RAG system.}
    \vspace{-5pt}
    \label{fig:system}
    \vspace{-5pt}
\end{figure}
\vspace{-5 pt}
\section{Related Work}
Recent work has shown that NLP methods can support policy analysis by helping researchers organize, interpret, and compare complex regulatory text \cite{van2021ai,papadopoulos2020governments,misuraca2020use,engstrom2020government}. Prior studies have applied text analysis to public policy and governance documents \cite{jin2022natural}, while legal and policy QA benchmarks such as PolicyQA \cite{ahmad2020policyqa}, PrivacyQA \cite{ravichander2019question}, and LegalBench-style tasks \cite{pipitone2024legalbench,askari2022expert,zhong2020jec} demonstrate that domain-specific corpora and task design are important for reliable assistance on regulatory text. However, most of this work does not focus specifically on the fast-moving and cross-jurisdictional domain of AI governance.

AI governance has recently emerged as a distinct application area for computational analysis. Large-scale resources such as AGORA collect AI-relevant laws, regulations, standards, and related metadata across jurisdictions, enabling systematic study of the regulatory landscape. 
Existing work in this space has largely emphasized corpus construction, descriptive analysis, and document-level assessment rather than interactive question answering grounded in primary policy text \cite{correa2023worldwide,Mavrogiorgos2023oced}. This leaves an important gap for systems that can answer targeted questions about provisions, definitions, authorities, and timelines across multiple policy documents.

LLMs have expanded the toolkit for high-stakes legal and policy text analysis \cite{blair2024blt,blair2023can,tang2023policygpt,bommarito2022gpt}, but prior evaluations show that they struggle with domain-specific terminology, nested definitions, and cross-references, and may hallucinate unsupported claims when used without grounding \cite{pipitone2024legalbench,dahl2024large,henderson2022pile}. RAG addresses this latter problem by conditioning generation on retrieved evidence, making it a natural approach for policy corpora. At the same time, the effectiveness of RAG depends on both retrieval quality and generation behavior. 
Recent work explores improving RAG through retriever training or preference-based alignment \cite{wu2024paragragalignmentmultiperspective}. We study these ideas in the context of AI policy corpora and analyze how improvements in retrieval interact with end-to-end question answering performance.

\section{Dataset}
Our system is built on the AGORA dataset, a curated collection of AI-related policy documents from multiple jurisdictions. The corpus contains $947$ policy documents (chunked into $7893$ chunks), including laws, regulations, standards, and policy guidelines, along with metadata such as issuing authority, policy sector, document type, enactment date, and thematic tags covering risks, harms, and policy interventions. We use the full-text policy documents provided by AGORA and index their pre-defined document segments as retrieval units, where each segment represents a coherent policy provision.
%
To support training and evaluation, we generate synthetic policy-related questions using document text and metadata fields. For generator alignment, pairs of responses across $2000$ questions grounded in the corresponding policy documents are compared across sample pairwise preference tasks to construct DPO training data. This process produces a dataset of document-grounded policy questions and answers suitable for both retriever training and generator alignment. Detailed statistics about document lengths, segment distributions, and metadata frequencies are provided in App. \ref{app:data}.
\section{Methodology and Experiments}
Our system follows a standard RAG architecture consisting of a retriever and a generator adapted to the AI policy domain (Fig. \ref{fig:system}).
\vspace{-5pt}
\paragraph{Retriever.}
We use ColBERTv2 as the base retriever and fine-tune it using contrastive learning. Synthetic queries are generated from AGORA documents using an LLM with prompts that incorporate metadata such as policy tags, authorities, and dates (Fig. \ref{fig:retriever-training}). For a training set of 200 of the queries, retrieved passages are manually labeled as relevant or irrelevant (relative to answering the query) to create training triples (query, positive passage, negative passage). The retriever is then optimized using a contrastive objective that increases similarity between queries and relevant passages while decreasing similarity to negatives.
\vspace{-5pt}
\paragraph{Generator alignment.}
The generator is based on Mistral-7B-Instruct and aligned using DPO. To construct preference data, we generate pairs of responses to the same policy question using different prompting and decoding strategies as mentioned in App. \ref{app:dpo-train-data-gen}. Human annotators select the preferred response given the document context, producing (prompt, chosen, rejected) training triples used for DPO fine-tuning, where we use parameter-efficient fine-tuning (PEFT) with LoRA \cite{hu2021loralowrankadaptationlarge} (Fig. \ref{fig:dpo-training} in App. \ref{app:method}).
\vspace{-5pt}
\paragraph{RAG pipeline.}
At inference time, user queries are encoded by the retriever to retrieve the top-k policy segments from the AGORA index. These retrieved passages are then provided to the generator together with the query to produce a grounded answer.

Methodological details are provided in App. \ref{app:method}.
\begin{table}[h]
    \centering
    \small
    \begin{tabular}{c|c|c}
        Retriever/Generator & Relevancy & Accuracy \\ \hline
        Base Mistral (w/o RAG) &0.733&0.581\\
        Base ColBERT/Base Mistral &0.746&0.605\\
        CL ColBERT/Base Mistral &0.744&0.580\\
        Base ColBERT/DPO Mistral &\textbf{0.747}&0.601\\
        CL ColBERT/DPO Mistral &0.700&0.559\\
        GPT-5.4 &0.744&\textbf{0.777}\\
        \hline
    \end{tabular}
    \caption{End-to-end QA performance.}
    \vspace{-5 pt}
    \label{tab:system}
    \vspace{-5 pt}
\end{table}
\begin{table*}
    \centering
    \small
    \begin{tabular}{c|c|c|c|c|c|c|c}
        Retriever & MRR & Recall@5 & Recall@10 & Recall@20 & MAP@5 & MAP@10 & MAP@20 \\
        \hline
        ColBERTv2-base & 0.641757 & 0.220146 & 0.347234 & \textbf{0.585305} & 0.456694 & 0.441440 &  0.481813 \\
        Mined negatives & \textbf{0.748333}
& 0.289443 & 0.384164 & 0.483517
 & \textbf{0.584472} & \textbf{0.561838} & \textbf{0.561443}
 \\
        Labeled negatives & 0.727095  & 0.267648 & 0.397876 &  0.506035 & 0.504367 &  0.517485 & 0.502683 \\
        Mixed negatives & 0.725857 & \textbf{0.300559} &  \textbf{0.401189} & 0.484138 & 0.528367 & 0.492509 & 0.476172 \\
    \end{tabular}
    \caption{Retriever performance.}
    \vspace{-5 pt}
    \label{tab:retriever}
    \vspace{-5 pt}
\end{table*}
\subsection{Evaluation}
We evaluate our system on both \textit{question answering} and \textit{retrieval} tasks over the AGORA corpus.
\vspace{-5pt}
\paragraph{Evaluation dataset.}
We construct an evaluation set of  $300$ policy-related questions derived from factual analyses of AI governance documents. Questions are created using statements from expert policy commentary and verified with assistance from domain experts.
\vspace{-5pt}
\paragraph{Evaluation metrics.}
For end-to-end QA evaluation, we measure answer relevance and answer accuracy following the RAGAS \cite{es2025ragasautomatedevaluationretrieval} evaluation framework. To assess generator grounding behavior, we also compute a faithfulness score that measures whether claims in generated responses are supported by retrieved context.

For retriever evaluation, we measure Mean Reciprocal Rank (MRR), Recall@k, and Mean Average Precision (MAP@k) on a labeled set of 50 queries with manually annotated relevant passages. Implementation details, training prompts, and hyperparameters are provided in App. \ref{app:exp}.

\section{Results and Analysis} 
Table \ref{tab:system} shows end-to-end QA performance across retriever and generator configurations. Domain-specific fine-tuning does not consistently improve performance over the baselines.
The best performing RAG configuration (Base ColBERT + DPO Mistral) only marginally improves answer accuracy compared to the baseline.
In contrast, the GPT-5.4 baseline achieves substantially higher answer accuracy despite not having access to web search, mainly due to sheer size difference (both in architecture and training corpora) along with more sophisticated post-training alignment for uncertainty calibration. These results suggest that improvements to individual RAG components do not necessarily translate into improved end-to-end performance in policy QA settings.

Table \ref{tab:retriever} reports retrieval performance for the baseline ColBERTv2 retriever and fine-tuned variants (described in App. \ref{app:example-labeling}). Overall, fine-tuning improves several retrieval metrics, including Mean Reciprocal Rank (MRR), Recall@5, and MAP@5. For example, the mined-negatives retriever achieves the highest MRR among all variants, while the mixed-negatives configuration achieves the best Recall@5. However, these improvements do not consistently translate to gains at larger retrieval depths (e.g., Recall@20), where the base retriever remains competitive.

We also evaluate generator \textit{faithfulness} by comparing the base Mistral model and the DPO-aligned variant. The DPO model achieves a faithfulness score of \textbf{0.80} compared to \textbf{0.78} for the base model, indicating a modest improvement in grounding.
\subsection{Expert Review}
To complement automatic evaluation, we conduct a small qualitative review with policy researchers familiar with AI governance in Turkey and the EU. Experts found that the system generally captures key policy themes and cited relevant document segments. However, they also identify limitations in generated answers, including incorrect assumptions about policy mechanisms (e.g., overstating risk-tiering in Turkey’s AI strategy), missing substantive policy details, and limited references to specific examples when invoking international standards. These observations indicate that while the system summarizes policy documents well, it still struggles with precise policy interpretation and cross-document grounding.
Additional examples are provided in App.~\ref{sec:expert_analysis}.


\subsection{Error Analysis}
We analyze representative failure cases to better understand system behavior.
\vspace{-5pt}
\paragraph{Missing documents in the corpus.}
Some evaluation questions reference recent policy developments that are not yet included in the AGORA corpus (examples are provided in App. \ref{app:sys_error}). In these cases, the retriever often returns related but outdated documents, which the generator may treat as relevant evidence. This can lead to confident but incorrect answers grounded in partially related policy text.
\vspace{-5pt}
\paragraph{Cross-jurisdiction retrieval errors.}
Because many AI policies deliberately share and reuse similar terminology, the retriever occasionally returns passages from the wrong jurisdiction. For example, questions about policies in one country may retrieve documents from other countries with similar regulatory language (App. \ref{app:err_country}). When such passages are provided as context, the generator may incorrectly attribute the retrieved content to the queried jurisdiction.
\vspace{-5pt}
\paragraph{Complex query interpretation.}
Some errors arise from queries involving multiple documents or nuanced constraints (e.g., comparing policies or excluding certain authorities). Details are in App. \ref{app:retrv_error}. In these cases, the retriever may prioritize semantically similar passages rather than those that satisfy all query conditions, leading to partially relevant but incorrect or incomplete retrieval results.

These observations explain why improvements in retrieval metrics do not necessarily translate to better end-to-end question answering performance.
\section{Conclusion} 
We present a domain-adapted RAG system for question answering over AI governance documents using the AGORA corpus. Our system combines retriever fine-tuning with contrastive learning and generator alignment via DPO. Experimental results show that while retriever fine-tuning improves retrieval metrics, these improvements do not consistently translate into better end-to-end question answering performance. In several cases, stronger retrieval led to more confident hallucinations when relevant documents were missing from the corpus. These findings highlight an important challenge for policy-focused RAG systems: optimizing individual components may not guarantee more reliable, grounded, or useful responses. Future work should explore stronger hallucination mitigation strategies, cross-document contextual grounding, and improved handling of document status changes.

\section{Limitations}
\subsection{Compute and model scale}
This work was conducted with limited computational resources, which minimized compute and energy usage but constrained the scale of models and experiments we were able to evaluate. In particular, the generator was based on a 7B-parameter instruction-tuned model rather than larger state-of-the-art LLMs. While this choice allows the system to be trained and deployed using relatively modest hardware, larger models may exhibit different behavior in policy-domain question answering, particularly with respect to hallucination mitigation and reasoning over complex regulatory text. Understanding strengths and weaknesses of different models for performance as well as logistical feasibility of day-to-day use is an important consideration in high-stakes tasks involving policy documents.

While we employ open-source LLMs to promote reproducibility and reduce cost barriers, these models may encode biases present in their training data \cite{islam2026gets,cuconasu2025rag,blodgett2020language,brown2020language}. We do not explicitly measure or mitigate such biases in this work.

\subsection{Evaluation coverage}
Our evaluation dataset was constructed from policy analyses and expert commentary on recent AI governance developments. Although this approach provides realistic policy-oriented questions, it also introduces a limitation: some questions reference policies that are not yet included in the AGORA corpus. When such cases occur, the retriever may return related but incomplete evidence, which can lead to incorrect grounded answers. While this reflects a realistic deployment scenario for evolving regulatory corpora, it may disadvantage RAG systems that rely strictly on coverage of pre-existing corpora.

\subsection{Preference data for alignment}
Generator alignment was performed using a relatively small set of pairwise preference annotations due to the limited availability of domain experts during initial development. As a result, the learned preference signal may not fully capture the expectations of policy researchers or practitioners. Larger and more diverse expert-labeled preference datasets could potentially improve the effectiveness of preference-based alignment for policy question answering.

\section{Ethical Considerations}
The AGORA dataset is open-source and constructed entirely from public records. We use AGORA in accordance with its license, i.e., for non-commercial use, and the RAG system is constructed with close cooperation with the AGORA team, including discussion over ethical issues. While the system retains several weaknesses common to language models (e.g., potential hallucinations and political biases), the research here is aimed at prototyping a tool, and we emphasize that such a tool should not be put into production for high-stakes settings like policy analysis until it passes sufficient quality checks. As a starting point, we emphasize transparency about design and capabilities, open-source data and code, and are evaluating responses with domain experts to guide development and develop usage guidelines.

\section{Acknowledgments}
We are thankful for the contributions of our talented team of rsearchers, including undergraduate and graduate students as well as faculty from the Department of Computer Science and Department of Political Science of Purdue University working through the \href{https://www.grail-lab.org/}{Governance and Responsible AI Lab}. As of March 2026, they included Mahule Roy, Ruth Sugiarto, Seunghyun Yoo, Nakshatra Tondepu, Tri Vo, Sarah Mohapatra, Selen Dogan Kosterit, and additional members of the broader AGORA team at Purdue and Georgetown University.

\bibliography{custom}

\appendix

\section{Dataset Details}
\label{app:data}
\begin{figure*}
    \centering
    
    \begin{subfigure}{0.32\textwidth}
        \centering
        \includegraphics[width=\linewidth]{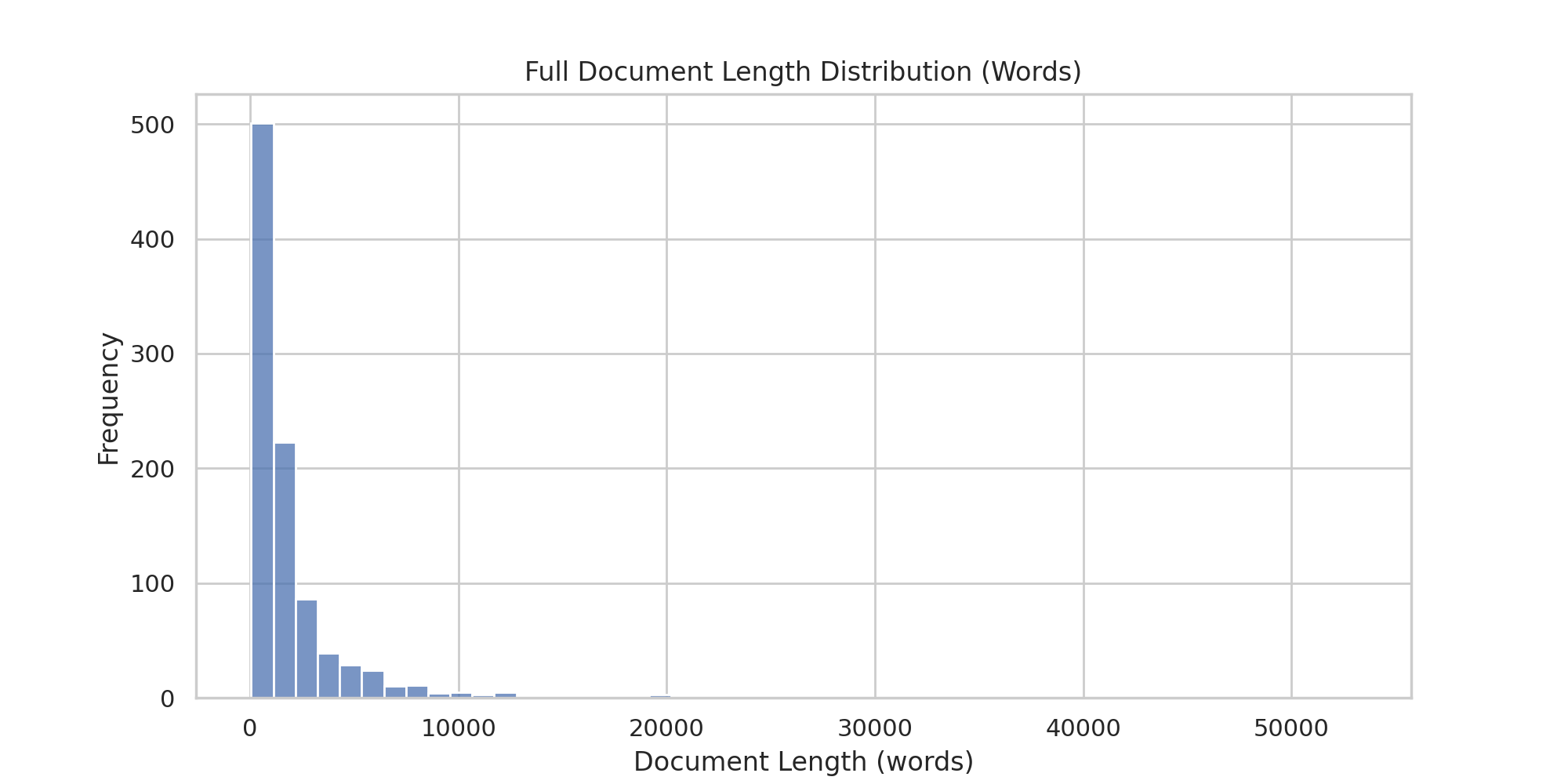}
    \caption{Document lengths}
    \label{fig:doc_length}
    \end{subfigure}
    \hfill
    \begin{subfigure}{0.32\textwidth}
        \centering
        \includegraphics[width=\linewidth]{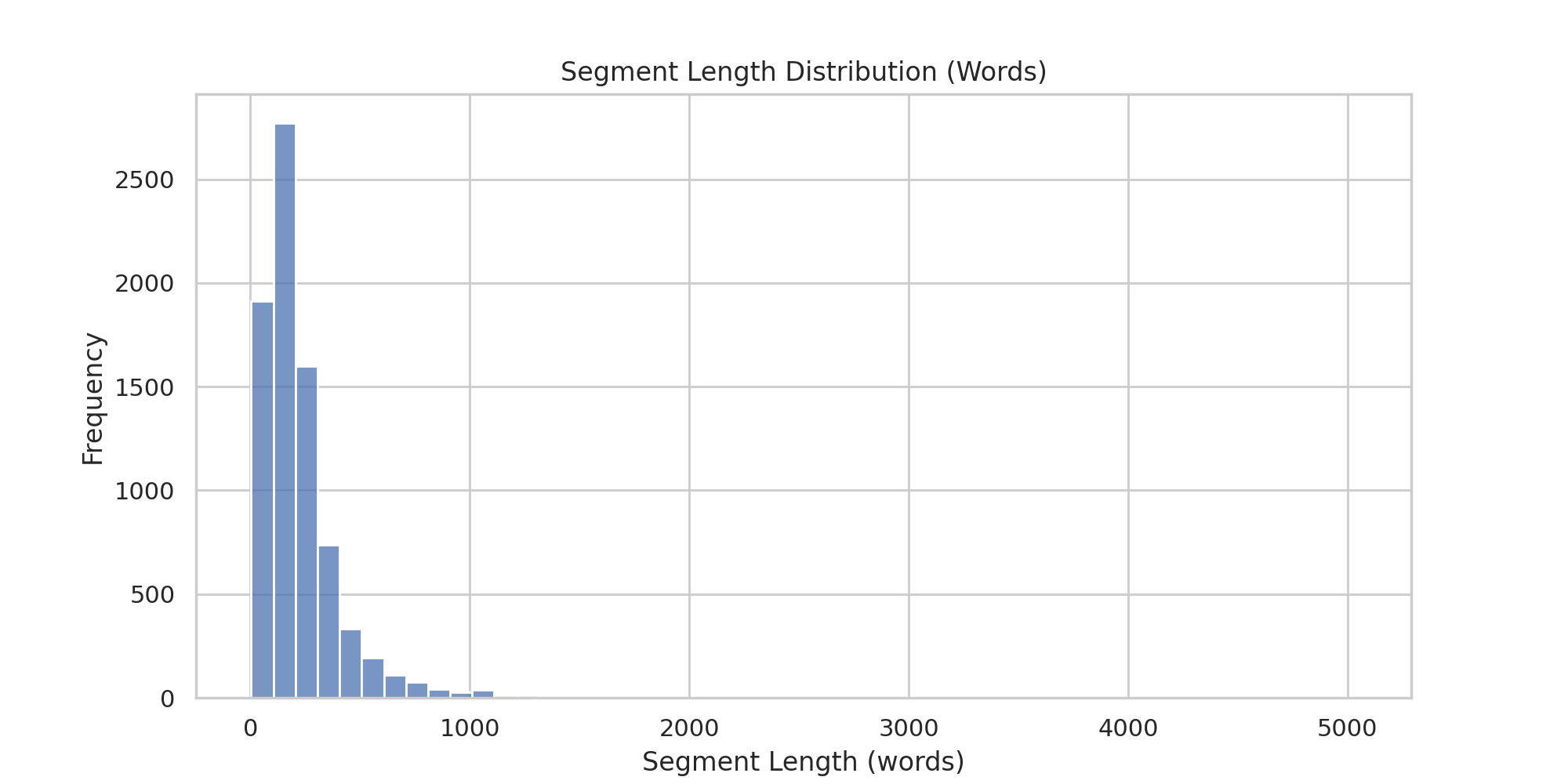}
    \caption{Segment lengths}
    \label{fig:seg_length}
    \end{subfigure}
    \hfill
    \begin{subfigure}{0.32\textwidth}
        \centering
        \includegraphics[width=\linewidth]{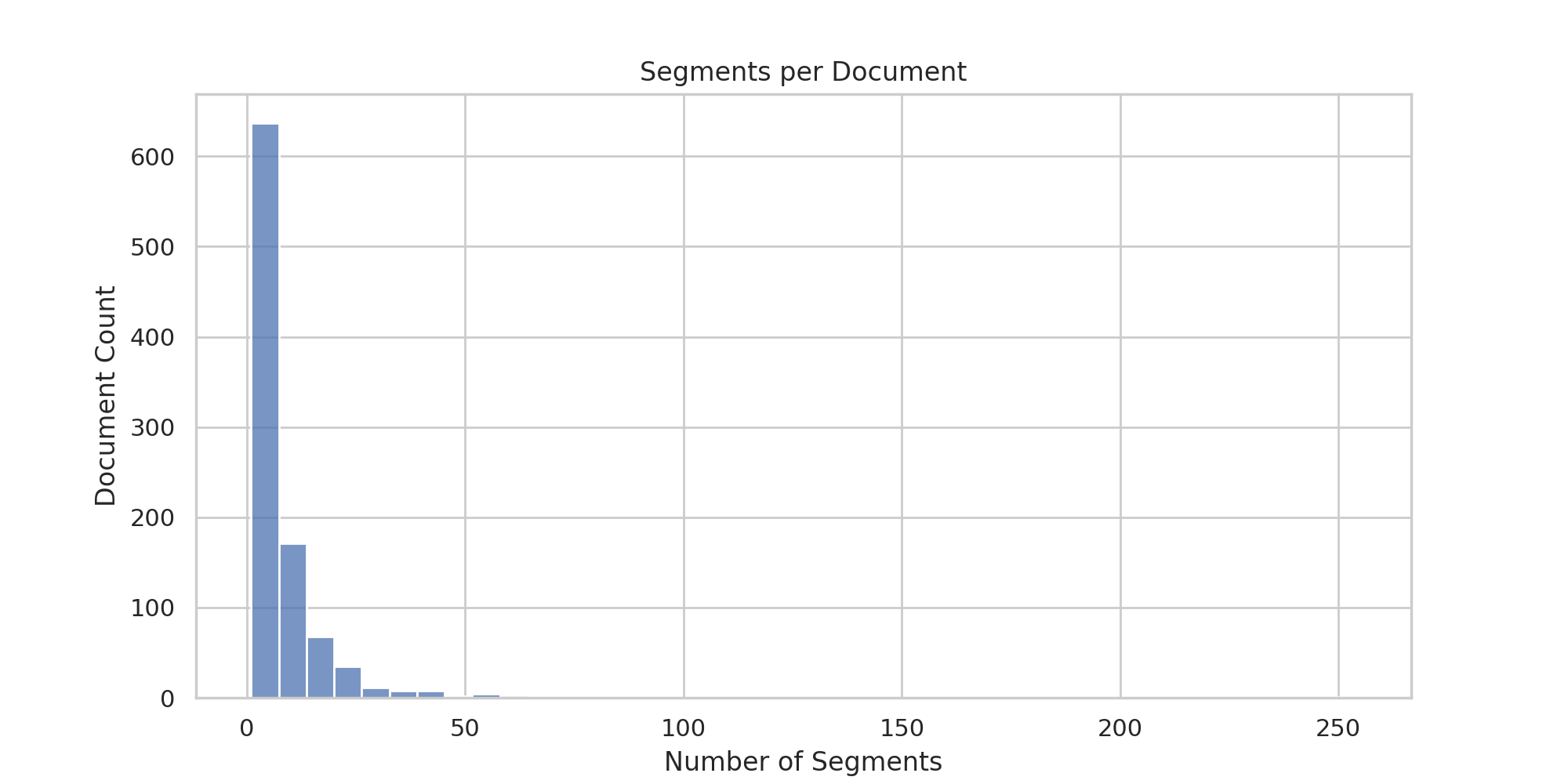}
    \caption{Segments per document}
    \label{fig:segs_per_doc}
    \end{subfigure}
    
    \vspace{0.5cm}
    
    \begin{subfigure}{0.32\textwidth}
        \centering
        \includegraphics[width=\linewidth]{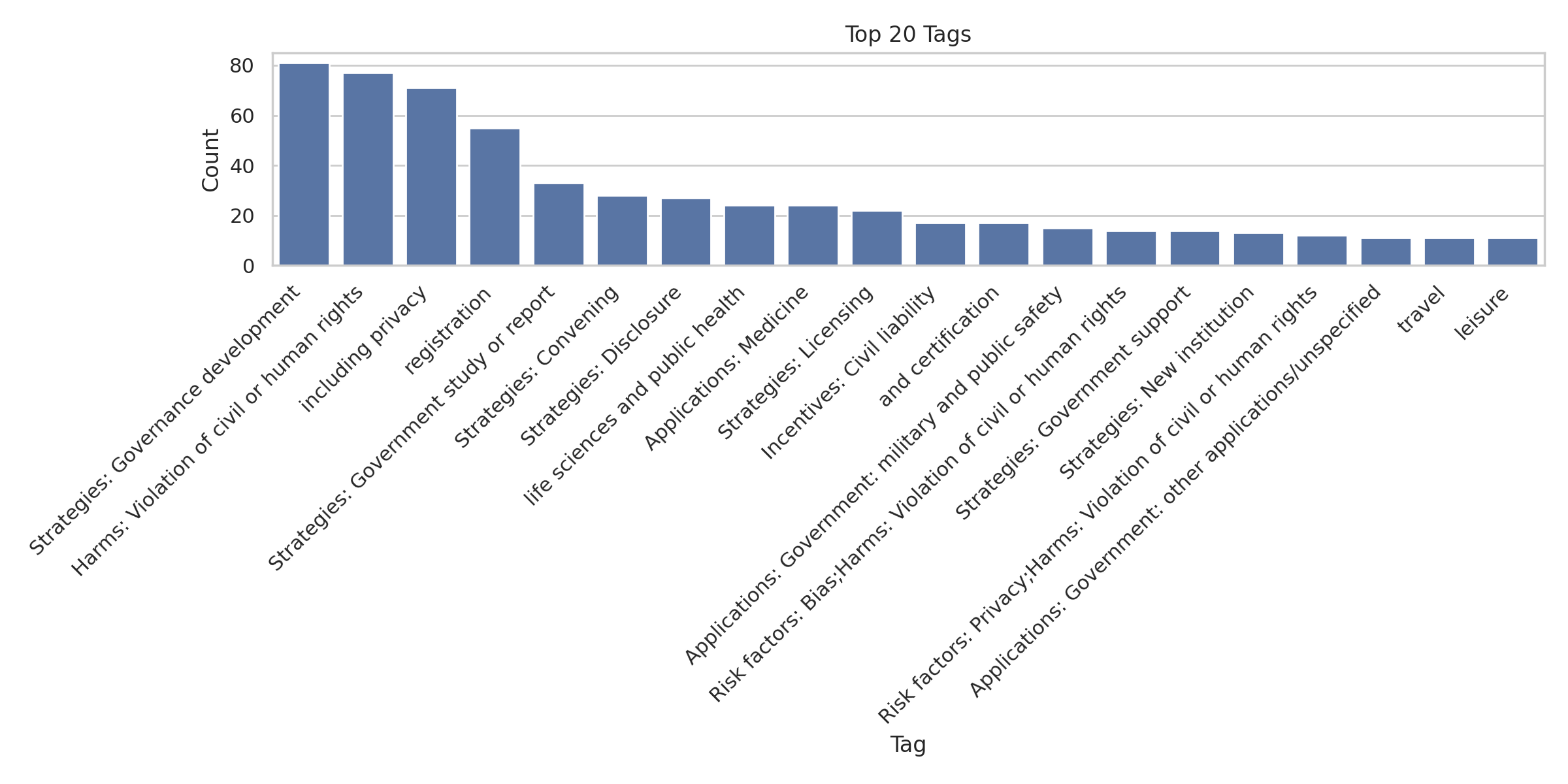}
    \caption{Top 20 tags}
    \label{fig:tags}
    \end{subfigure}
    \hfill
    \begin{subfigure}{0.32\textwidth}
        \centering
        \includegraphics[width=\linewidth]{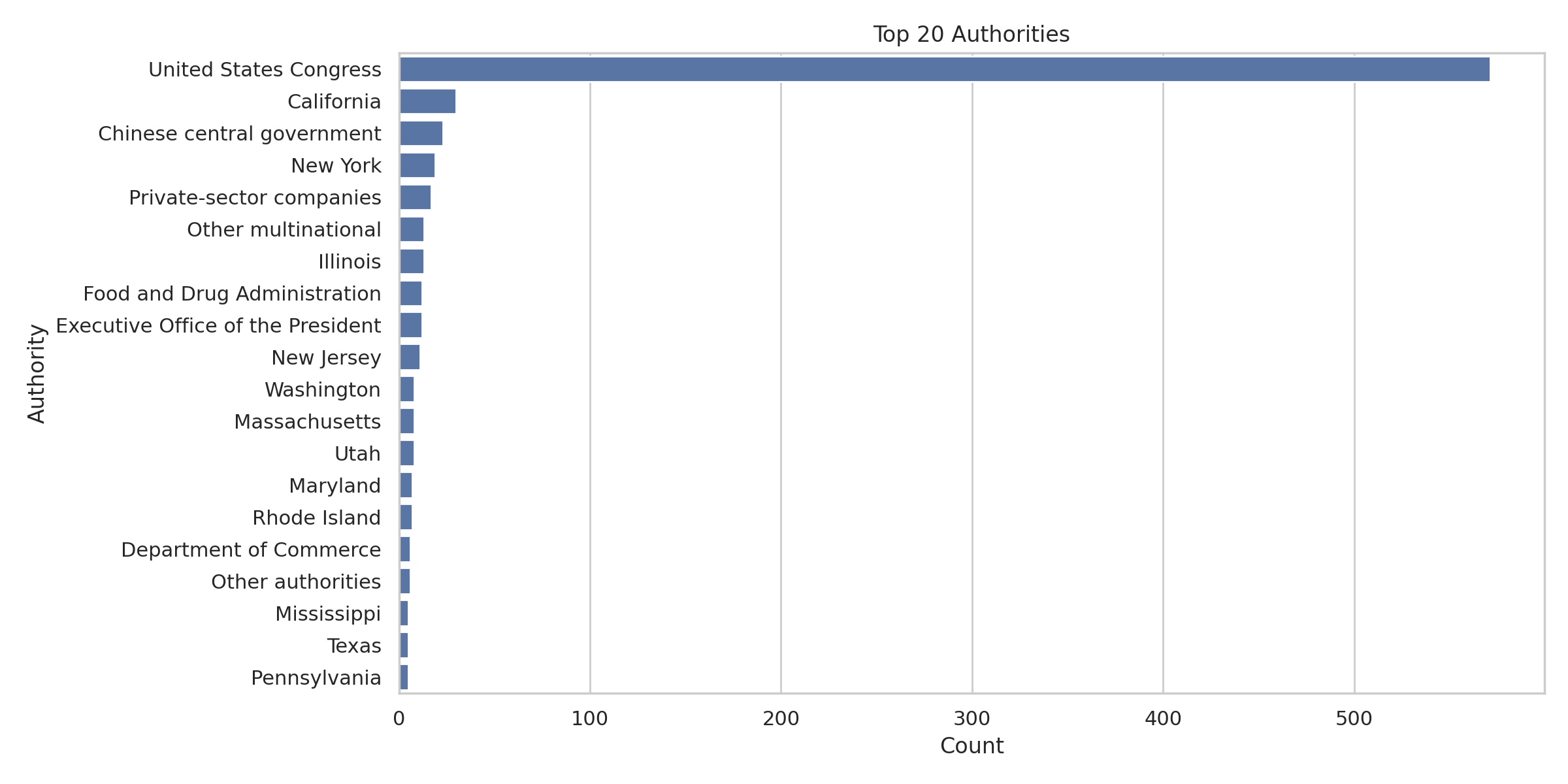}
    \caption{Top 20 authorities}
    \label{fig:auths}
    \end{subfigure}
    \hfill
    \begin{subfigure}{0.32\textwidth}
        \centering
        \includegraphics[width=\linewidth]{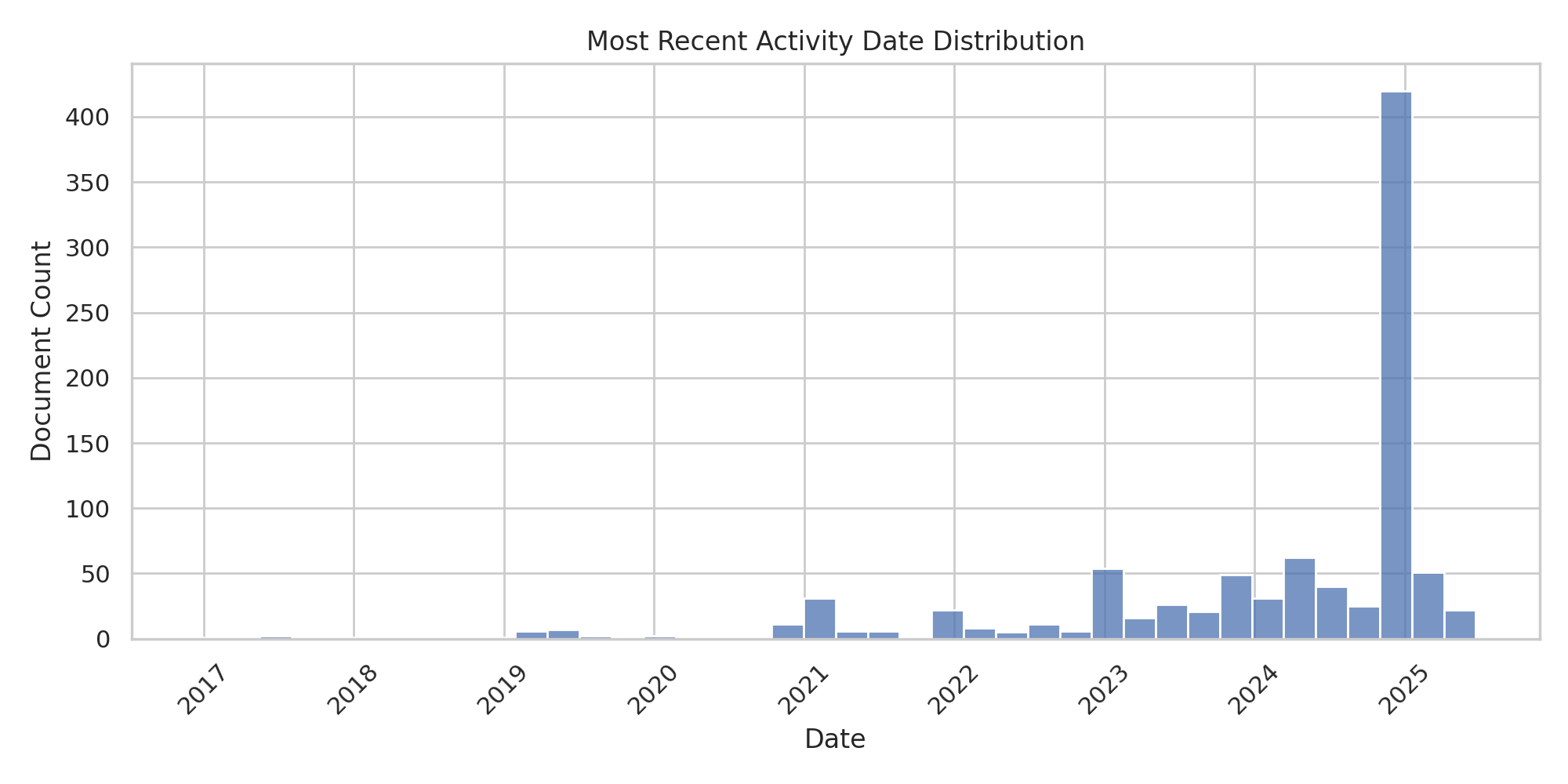}
    \caption{Dates of documents} 
    \label{fig:dates}
    \end{subfigure}  
    \caption{Data distribution.}
\end{figure*}

We perform data analysis to quantify the size of documents, the size of segments, the number of segments per document, and the frequencies of the tags, authorities and dates present in the dataset. We find that documents are almost all shorter than 5000 words, but some are much longer (Fig. \ref{fig:doc_length}). The vast majority of segments are less than 400 words with an average length of 226 words, and more than 99\% are less than 1000 words (Fig. \ref{fig:seg_length}). There are on average 8 segments per document, and more than 99\% of documents have less than 50 (Fig. \ref{fig:segs_per_doc}). There are 1702 unique tags in AGORA, with only 10 appearing more than 20 times (Fig. \ref{fig:tags}). By far the most common authority is the US Congress, being the authority of more than half of all documents in AGORA (Fig. \ref{fig:auths}). The most recent activity for all documents is no earlier than 2017, and the vast majority are from the last 3 years with a large spike at the beginning of 2025 (Fig. \ref{fig:dates}).


\section{Methodological Details}
\label{app:method}
This paper proposes a RAG system constructed by fine-tuning both the retriever and the generator for AI governance and policy.
\subsection{Chunking}
\begin{figure}[h]
    \centering
    \includegraphics[width=\linewidth]{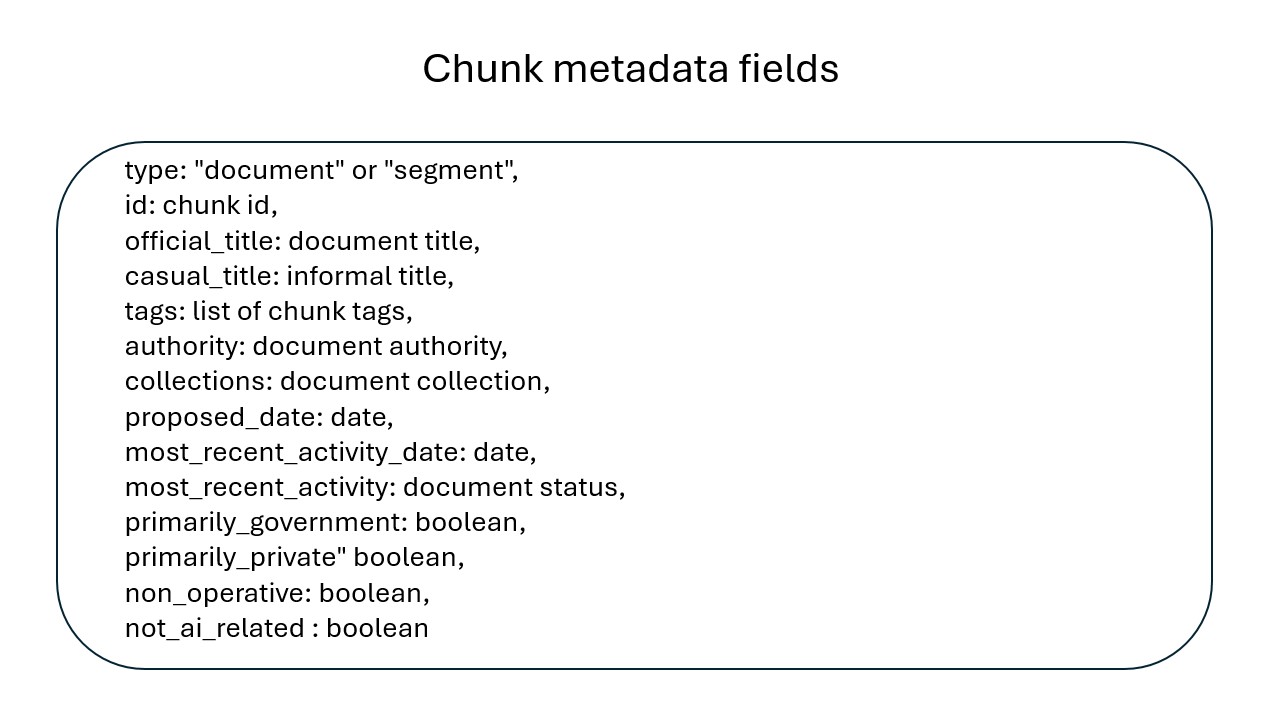}
    \caption{Chunk metadata.}
    \label{fig:chunk_format}
\end{figure}

As the dataset has already been manually chunked by AI policy researchers into segments, we simply use the segments as chunks. The segments are already constructed to chunk each policy instrument into relatively short sections that logically divide the document. The text from the chunks is stored with various annotations and metadata attached for the retriever to also see. This allows the retriever to accurately retrieve chunks relating to questions referencing things like tags, dates, document names, and authorities. The format of a chunk can be found in Fig. \ref{fig:chunk_format}. The "tags" field is only present if the segment has been annotated and labeled with one or more tags. 
\subsection{DPO Fine-Tuning}
\begin{figure*}
    \centering
    \includegraphics[width=1\linewidth]{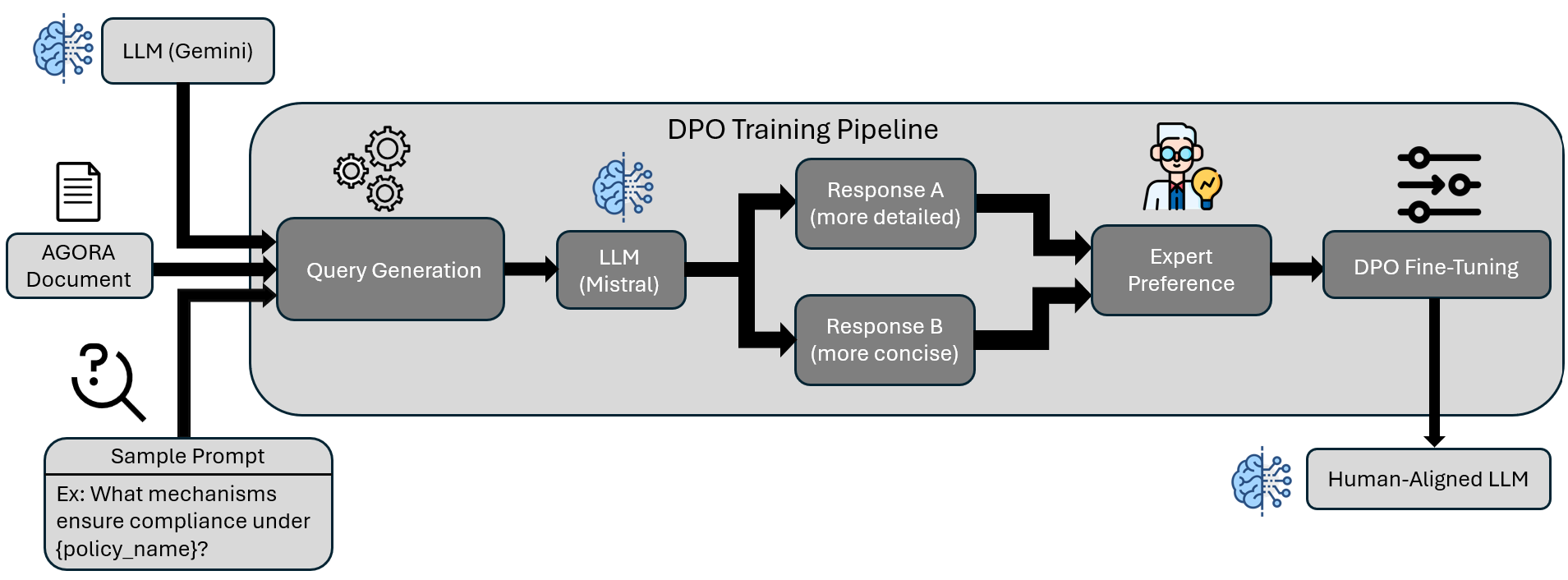}
    \caption{Generator DPO fine-tuning pipeline.}
    \label{fig:dpo-training}
\end{figure*}

Given a prompt $x$, a preferred answer $y^+$, and a rejected answer $y^-$, DPO directly adjusts the model to increase the relative likelihood of the preferred answer. The DPO objective is given by: 
\begin{equation}
\begin{aligned}
\mathcal{L}_{\text{DPO}} = &-\log \sigma\Big(
\beta \Big[
\log \pi_\theta(y^{+} \mid x) - \log \pi_\theta(y^{-} \mid x) \\
&- \log \pi_{\text{ref}}(y^{+} \mid x) + \log \pi_{\text{ref}}(y^{-} \mid x)
\Big] \Big)
\end{aligned}
\end{equation}
where $\pi_\theta$ is the fine-tuned policy model, $\pi_{\text{ref}}$ is a frozen reference copy of the base model, $\beta$ controls how far the model is allowed to deviate from the reference, and $\sigma$ is the sigmoid function. In Fig. \ref{fig:dpo-training}, after question/answer generation and collection of preferences, the base LLM (Mistral-7b-Instruct-v0.3) is trained according to the objective above.
\subsubsection{Training Data Generation }\label{app:dpo-train-data-gen} Questions for the training data are generated document-wise from the AGORA dataset, where we prompt an LLM to create a few questions based on that document, a given category, and an example format for that theme. Because the retrieval capabilities of the RAG system are not tested, we choose to provide the model with the reference document when it was generating the answer pairs for the questions.

To generate the training data for DPO fine-tuning, two distinct model configurations are used to produce responses of differing styles and quality. Both configurations are based on the same underlying model, but vary in their prompting and decoding strategies to promote diversity in outputs. The first configuration is designed to produce detailed, well-reasoned, and comprehensive responses grounded in the provided context. Its prompt positions the model as an "expert policy analyst" expected to deliver structured explanations and incorporate multiple perspectives. In contrast, the second configuration emphasizes brevity and clarity, encouraging concise and direct answers with minimal elaboration. This setup aims to capture more succinct, readable responses that might sacrifice depth for precision.
\begin{figure*}
    \centering
    \includegraphics[width=1\textwidth]{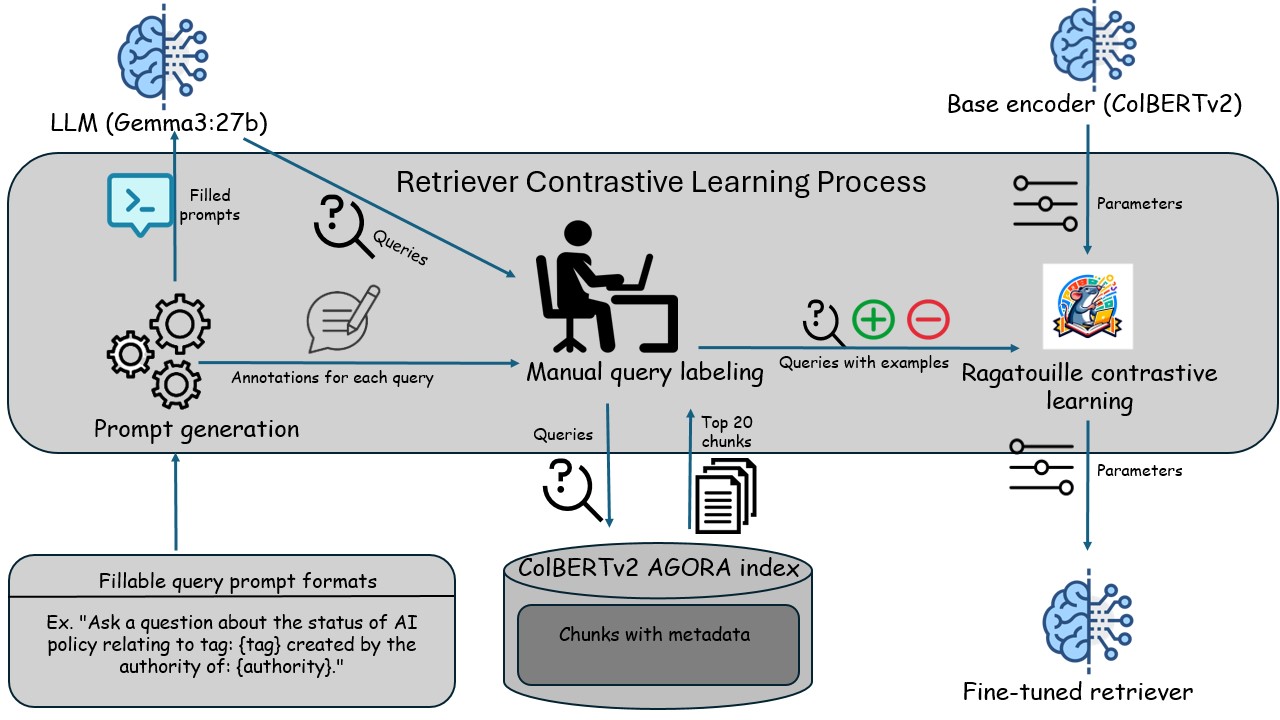}
    \caption{Retriever training pipeline.}
    \vspace{-5pt}
    \label{fig:retriever-training}
    \vspace{-5pt}
\end{figure*}
\subsubsection{Fine-Tuning Process}
    After we store the pairwise preferences for each model response, we use those to align our LLM with DPO. Each preference is stored as a triplet (prompt, chosen, rejected), where the prompt used is the one that encouraged detailed answers, including the document context. For computational efficiency, we load the Mistral-7b-Instruct model in an 8-bit quantization, which helps with GPU memory limitations during training.
    To run DPO efficiently on a single GPU, we use parameter-efficient fine-tuning (PEFT) with LoRA \cite{hu2021loralowrankadaptationlarge} adapters, meaning we only have to update a small number of low-rank matrices rather than the full model parameters. Furthermore, we use a gradient accumulation of 8 steps with a batch size of 2, simulating a batch size of 16.
    We find that $1$ epoch is a sufficient training time, as training for longer over a smaller preference dataset, as we have in our case, can lead to overfitting of annotation noise rather than improving alignment.
\subsection{Retriever Fine-tuning} \label{4.3.1}
As visualized in Fig. \ref{fig:retriever-training}, the retriever fine-tuning process is a pipeline of multiple steps to optimize a retriever for the AGORA dataset in an automatic process not requiring manual intervention. 
\subsubsection{Synthetic Query Generation}
For both the issue of evaluating a retriever as well as fine-tuning one, a set of queries with known relevant documents is necessary. As manually constructing a large set of these labeled queries requires significant time and effort, synthetic query generation is an efficient and effective method to create such a dataset \cite{wen2025syntheticdatastrategiesdomainspecific, kim2025syntrievertrainretrieversynthetic}. We construct a synthetic query generation pipeline specifically for retriever training and evaluation that creates queries with some relevant coverage over the range of possible user queries, and manually labeled a set of these queries with relevant and irrelevant documents. The question generation consists of a prompt generation system that creates thousands of prompts for an LLM to create queries for the retriever. Considering the research-focused use case of our system, we create prompts focused on analysis-related topics like trends and comparisons. With the AGORA dataset already being annotated at the document and chunk level with things like tags, authorities, and dates, combinations of those annotations are inserted into the prompts to create full prompts for the LLM that cover many possible topics within the scope of AI governance and regulation. These filled prompts are then passed to the LLM with instructions to create a question based on the prompt. 
\subsubsection{Positive/Negative Example Labeling}\label{app:example-labeling}
To fine-tune ColBERTv2, we need to label the queries with positive and negative examples from the chunks in the dataset. We discard queries if they are not well-formed or just generally not relevant. With questions that we have not discarded, we use ColBERTv2 to retrieve the top-20 chunks for each question, and manually label each chunk as \textit{relevant} or \textit{irrelevant}. This obviously gives no guarantee of finding the best positive and negative examples, but by manually labeling, we ensure that the training moves the retriever in the right direction. After this labeling process, (query, positive example, negative example) triples are created by creating a triple from each possible pair of positive and negative examples from the labeled sets of each query. For this reason, despite only $127$ queries being labeled, $8339$ training triples are created. Furthermore, the Ragatouille library makes available a hard negative mining feature that finds likely useful negative examples for any given query. This gave us three options for providing negative examples to the fine-tuning procedure. First, just using our labeled negatives and not using the mining procedure at all, second, only using the mined negatives and disregarding our labeled negatives, and third, using both the mined and labeled negatives. We fine-tune a retriever using each of these three methods. We refer to these fine-tuned retriever variants as ``Labeled negatives", ``Mined negatives", and ``Mixed negatives".
\subsubsection{Fine-tuning Procedure}
ColBERTv2 based retriever is fine-tuned using our synthetically generated and manually labeled query dataset using the Ragatouille libraries' RAGTrainer feature. The RAGTrainer performs a contrastive learning procedure to update the parameters of the ColBERTv2 retriever to make the embedding of the query closer to the embedding for the positive example and further from the negative example. It does this by optimizing a contrastive InfoNCE objective (2) that increases the ColBERT similarity 
$S(q,p)$ (1) between a query and its positive passage while decreasing similarity to negatives. 
\begin{align}
S(q,p) &= \sum_{t} \max_{s} \, \mathrm{sim}(u_t, v_s), \\
\mathcal{L}(q,p^{+},p^{-})
    &= - \log
       \frac{e^{S(q,p^{+})/\tau}}
            {e^{S(q,p^{+})/\tau}+e^{S(q,p^{-})/\tau}}.
\end{align}
In these equations, $q$ denotes a query and $p$ a passage. The query and passage are encoded into token-level embeddings, where $u_t$ is the embedding of the $t$-th query token and $v_s$ is the embedding of the $s$-th passage token. The similarity function sim$(u_t,v_s)$ measures the similarity between token embeddings. For training, the $\mathcal{L}(g,p^+,p^-)$ contrasts a relevant (positive) passage $p^+$ against an irrelevant (negative) passage $p^-$, using a temperature parameter $\tau$ to scale the logits. 

\subsection{RAG pipeline}
The system itself is a fairly standard RAG pipeline using the fine-tuned retriever and generator. The AGORA dataset is chunked simply by using the document segments as chunks. The segments are manually created by researchers when documents are added to the dataset to split the documents into semantically relevant chunks of a reasonable length, so simply using the segments as our chunks is the most logical decision. The chunks are encoded into an index using our fine-tuned retriever using the Ragatouille library's RAGPretrainedModel.index function. When running, the system receives user queries, encodes them using the fine-tuned retriever, and then using RAGPretrainedModel.search finds the top-20 chunks from the index. These chunks are passed to the fine-tuned generator along with the user query, and the generator responds to the user query.
\begin{figure*}
    \centering
    \includegraphics[width=1\linewidth]{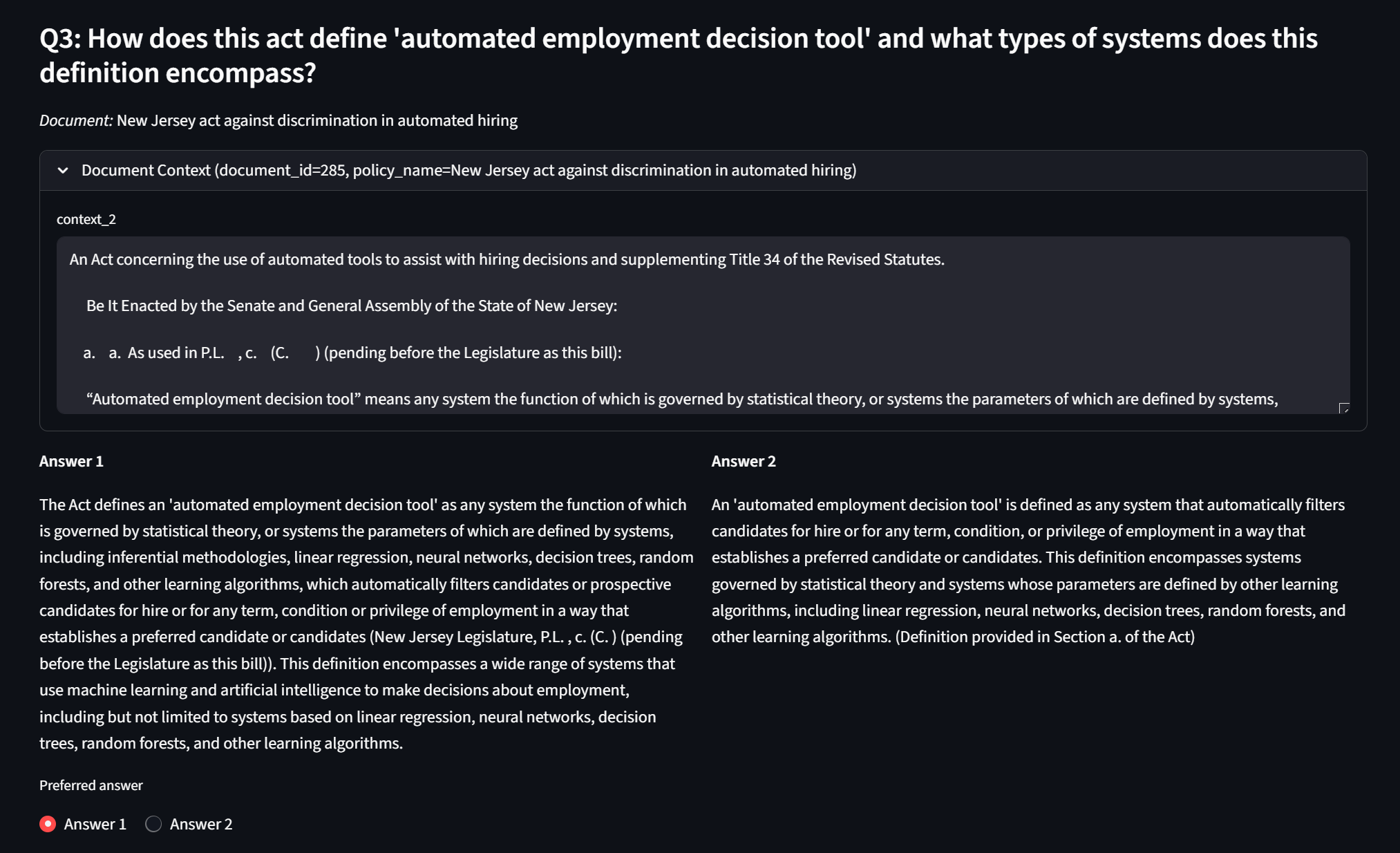}
    \caption{DPO response collection GUI.}
    \label{fig:dpo-gui}
\end{figure*}

\section{Experimental Details}
\label{app:exp}
\subsection{Evaluation Question Creation}
To evaluate our system, we need a set of relevant questions labeled with factually accurate answers. Based on the size of our document corpus, we determined that a set of $300$ questions and answers is a suitable size. We create the questions and answers from factual analysis of AI policy found in the AI Policy Corner\footnote{\url{https://montrealethics.ai/category/insights/ai-policy-corner/}} blog of the Montreal AI Ethics Institute. These articles feature analysis discussing the content and implications of recent AI policies from around the world, and comparing the approaches taken through different policies. Our process for creating an evaluation dataset for our system involved taking factual statements from these articles and creating questions about AI policy that those factual statements accurately answer. This is done both manually and with assistance from LLMs. In this process, we are assisted by contributions from policy experts involved in the writing of the articles.
\subsection{Generator}
\subsubsection{Generator Optimization}  
\begin{figure*}
    \centering
    \includegraphics[width=0.8\linewidth]{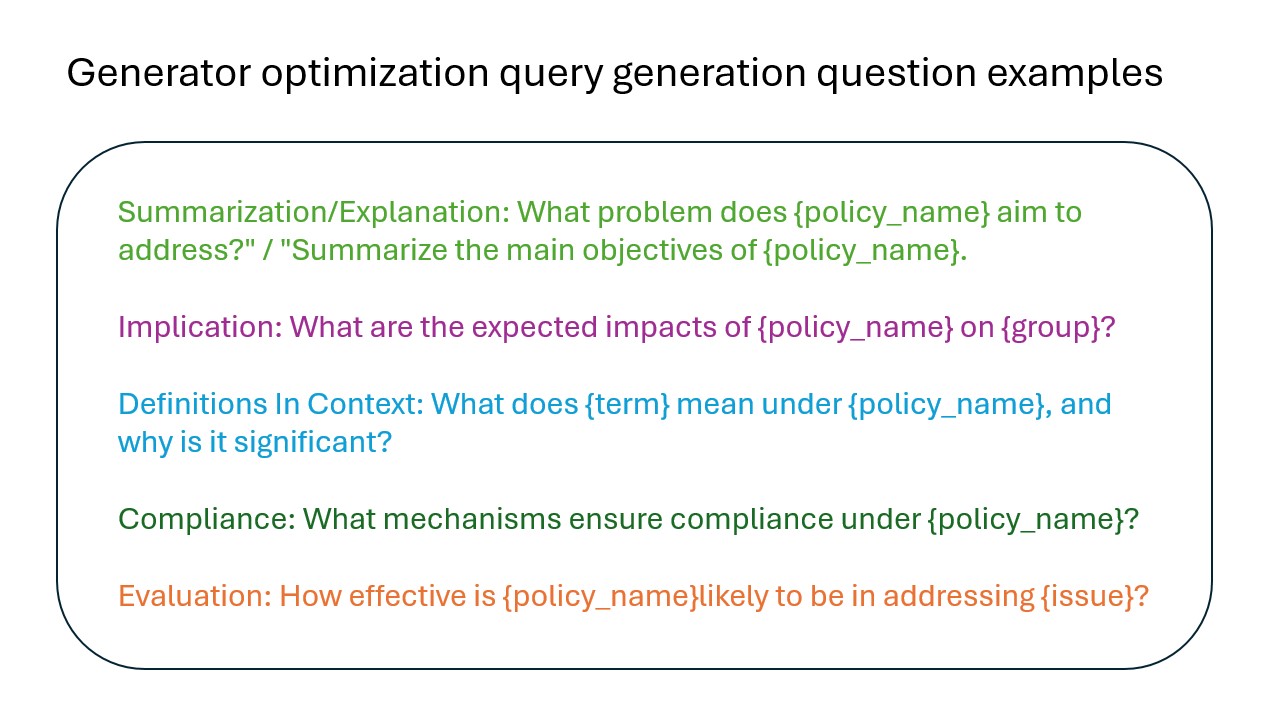}
    \label{fig:gen_qs}
    \includegraphics[width=0.8\linewidth]{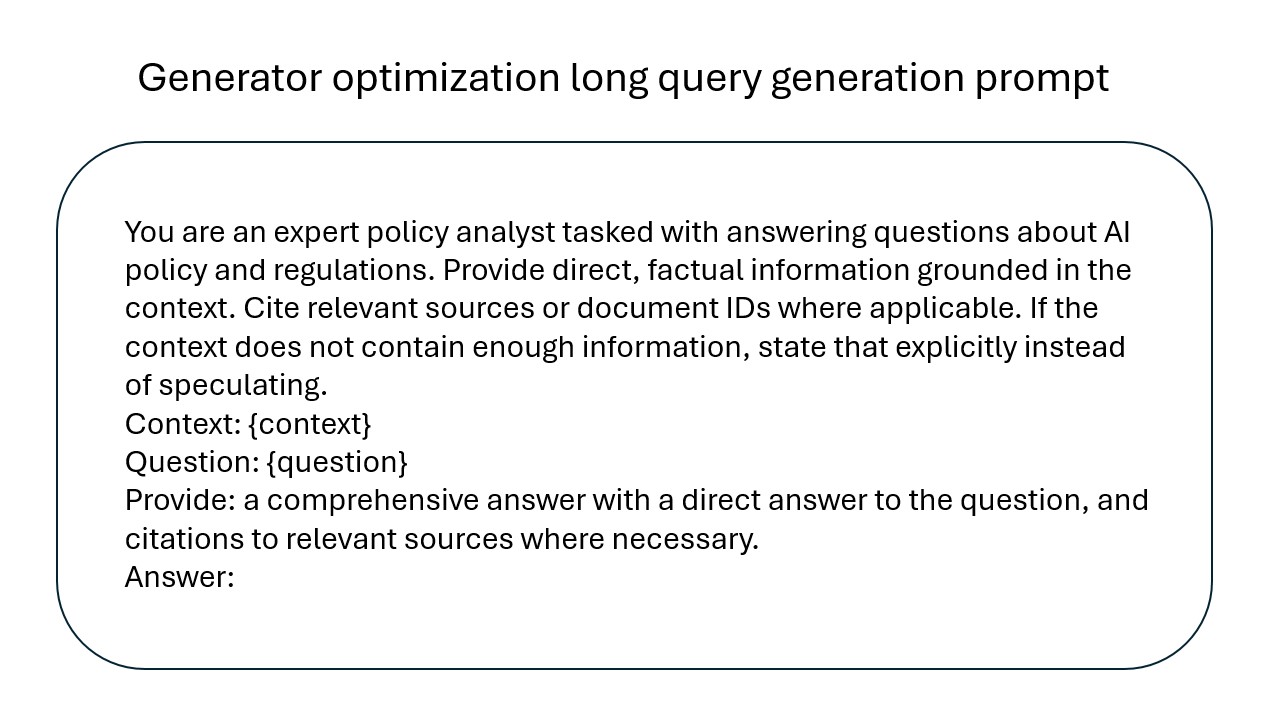}
    \label{fig:gen_long}
    \includegraphics[width=0.8\linewidth]{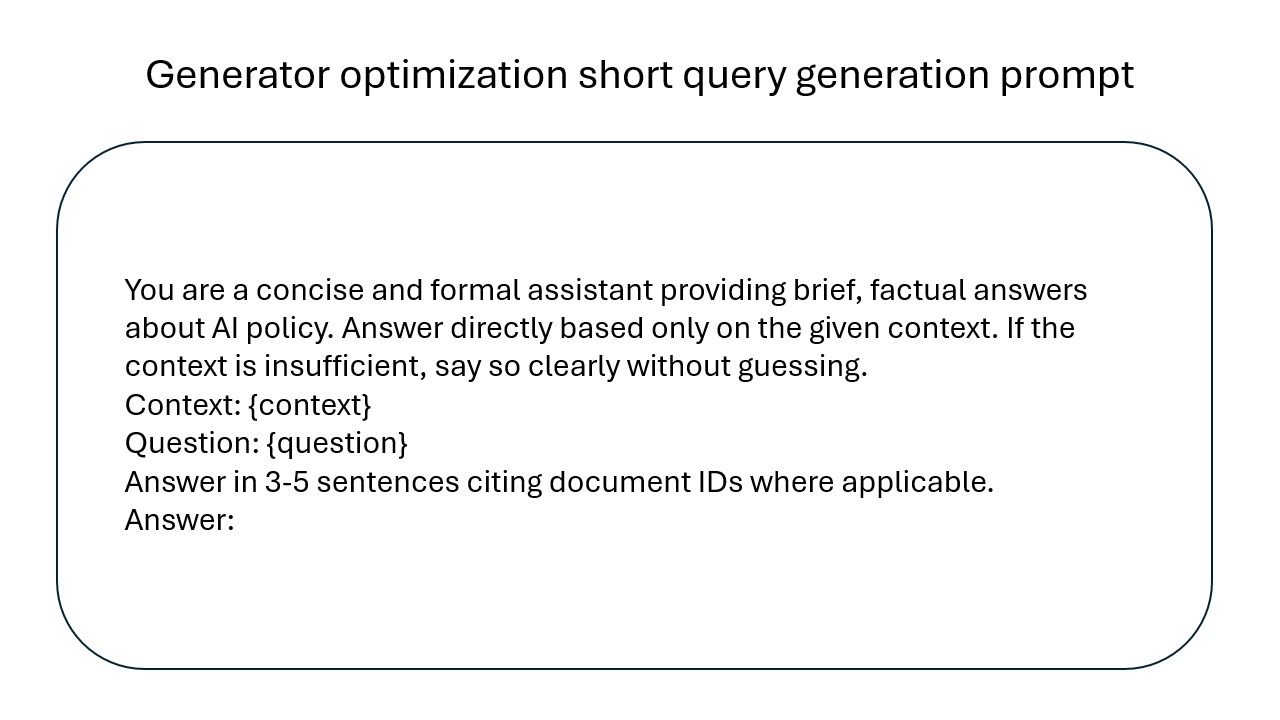}
    \label{fig:gen_short}
    \caption{Generator optimization prompts.}
    \label{fig:generator_prompts}
\end{figure*}

To generate the set of questions for the DPO training data, the gemini-2.0-flash and gemini-2.5-flash-lite models are used. For the documents, to avoid overloading the context length, only documents with word counts between 300 and 1200 words are chosen. The six categories that each question fell under are Summarization/Explanation, Implication, Stakeholder Interpretation, Definitions in Context, Compliance, and Evaluation, with each category having 299, 305, 349, 342, 368, and 337 questions, respectively, making a total of 2000 questions. In the prompt provided to the model to make the questions, the same general structure is preserved across categories, and during generation, along with specifying the category, an example question is added for the model to reference. Example questions provided to the model for each category are shown in Fig. \ref{fig:generator_prompts}

When actually generating the answer pairs for the questions, Mistral-7B-Instruct-v0.3 is used, which is the same model that was fine-tuned. The two different model configurations consisted of a different system prompt, as well as different choices for the \textit{temperature}, \textit{top\_p}, and \textit{top\_k} hyperparameters. The first configuration used \textit{temperature = 0.2, top\_p = 0.95}, and \textit{top\_k = 40}, paired with a prompt that emphasized detailed information grounded in the provided context. With the lower temperature and the higher top\_k, we intended this model configuration to generate in-depth explanations while still considering multiple points of view. The second configuration used \textit{temperature = 0.9, top\_p = 0.6}, and \textit{top\_k = 20}, along with a prompt that instructed the model to create brief responses to have more readable responses. The prompts used both for adding more detail and being more concise are shown in Fig. \ref{fig:generator_prompts}

After the generation of these answers, human preferences are collected through a streamlit interface as seen in Fig. \ref{fig:dpo-gui}, where the user can see the document context along with the questions and answers.
\subsubsection{DPO Training Loop}  We load the base model, Milstral-7b-Instruct-v2, in 8-bit quantized mode through the bitsandbytes Python library. Because fully quantized models cannot be fine-tuned directly, we added LoRA adapters with the PEFT library. This setup matches the QLoRA \cite{dettmers2023qloraefficientfinetuningquantized} configuration, where the 8-bit backbone remains frozen and only the LoRA layers (in FP16) are trainable. This training paradigm achieves a lot of memory savings, allowing us to fine-tune on a single A100 40GB GPU.

We use HuggingFace TRL's \cite{vonwerra2022trl} DPOTrainer, which automatically created a frozen reference model, handled the DPO objective along with the training loop, batching, and logging. As for hyperparameters, using a batch size of 2 with a gradient accumulation value of 8 led to an effective batch size of 16; the learning rate was $5 \times 10^{-6}$; we trained for 1 epoch, the $\beta$ value was 0.1; and the optimizer was paged\_adamw\_32bit, a memory-efficient optimizer for quantized models.

\subsubsection{Generator Evaluation}  To evaluate the generator without relying on ground-truth answers, we rely on one key metric: \textbf{faithfulness}. Faithfulness evaluates whether the generated response is supported by the retrieved context by decomposing the answer into individual claims and verifying that each is grounded in the provided evidence. With this metric, we essentially evaluate the generator's ability not to hallucinate and be truthful. The evaluation was conducted using RAGAS, an evaluation framework for RAG systems. The test set of $208$ questions was generated using the almost exact methodology as the training set for DPO; the only difference was using longer document lengths (1200+ words) so that there would be no overlap between the documents used for the training data (300-1200 words). For question categories, the same categories as the training data are used, and questions of each type are randomly distributed. After responses from the candidate models are collected, it is passed into the RAGAS faithfulness metric, where an LLM is used to extract all claims in the answer and verify what percent of those claims can be inferred from the given context. The LLM used to evaluate the responses is also Mistral-7B-Instruct-v0.3.
\subsection{Retriever}
\subsubsection{Synthetic Query Generation}
\begin{figure*}
    \centering
    \includegraphics[width=0.8\linewidth]{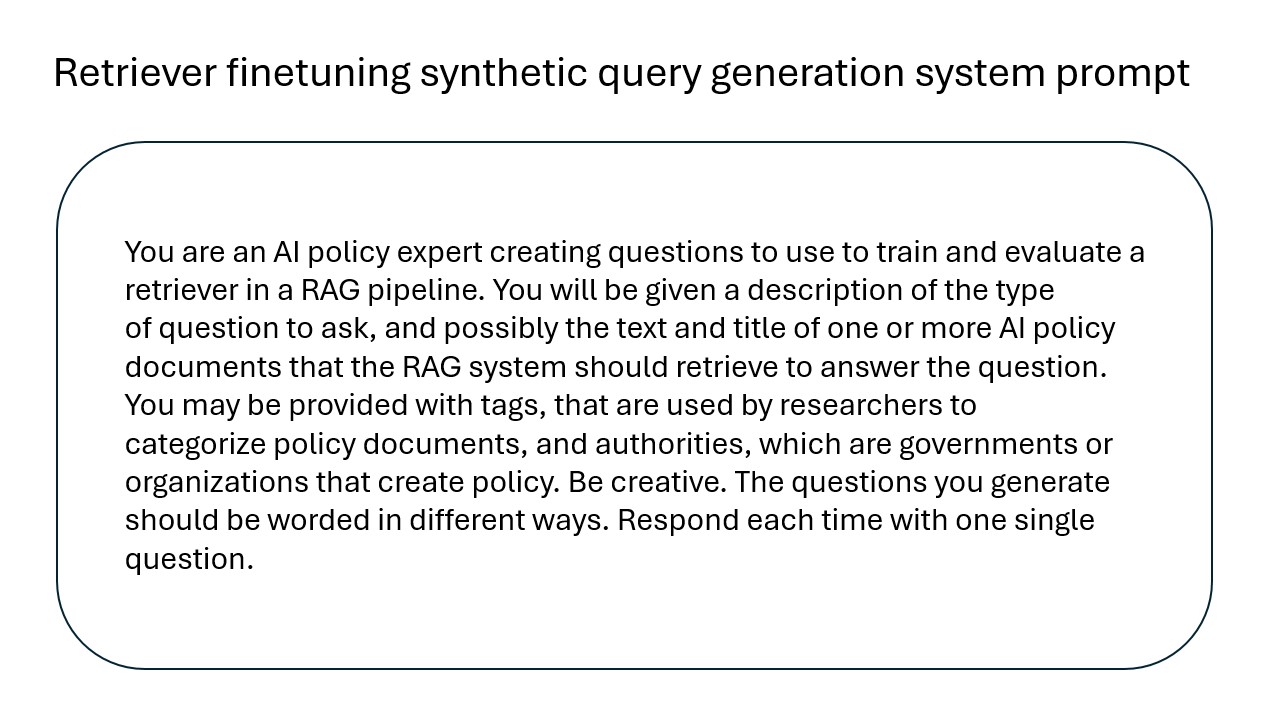}
    \label{fig:ret_sys}
    \includegraphics[width=0.8\linewidth]{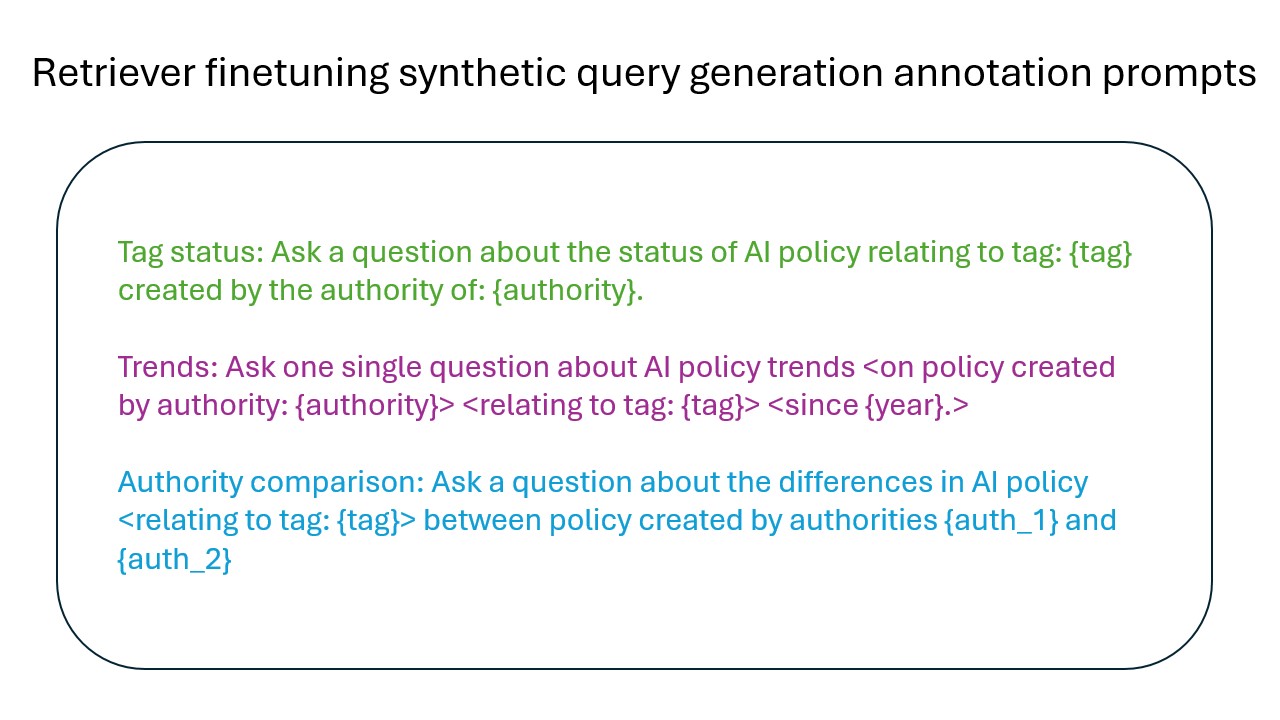}
    \label{fig:ret_annot}
    \includegraphics[width=0.8\linewidth]{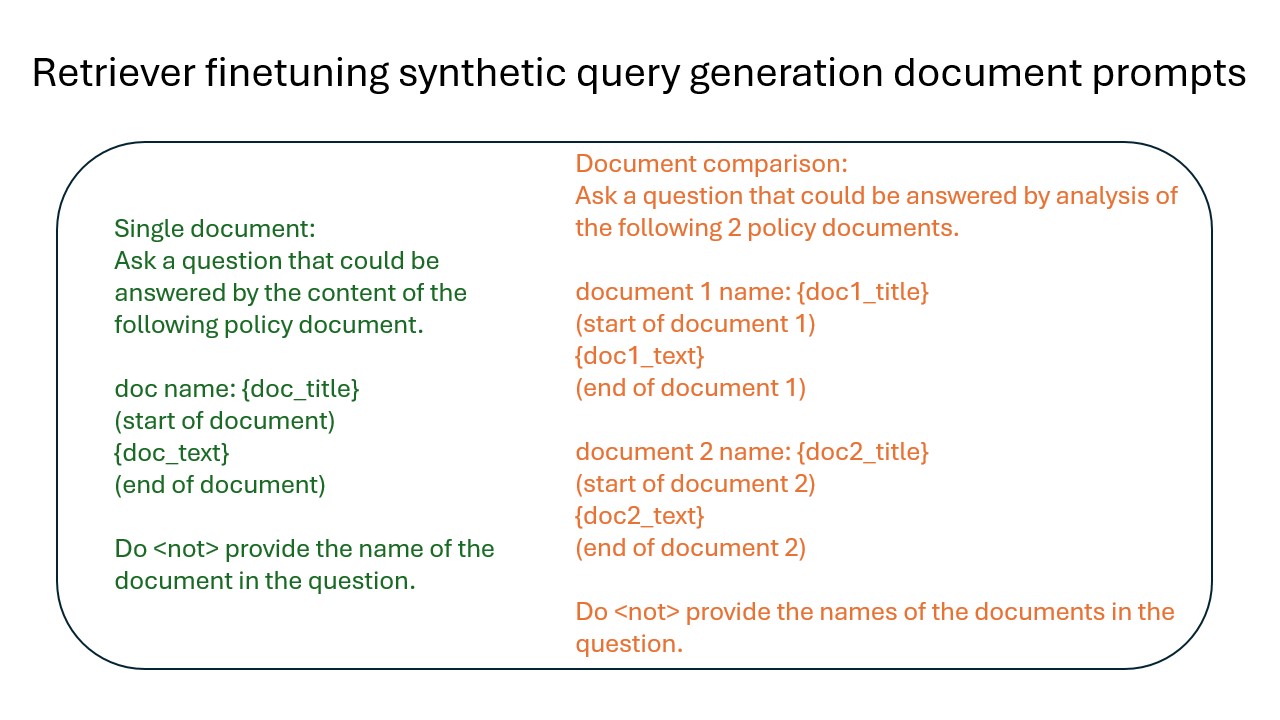} 
    \label{fig:ret_doc}
    \caption{Retriever fine-tuning prompts.}
    \label{fig:retriever_figs}
\end{figure*}

As discussed in \ref{4.3.1}, we need to synthetically generate queries to use to both train and evaluate the retriever. To do this, we use gemma3:27b to generate questions based on a set of prompts. The prompts contain places for tags, authorities, and dates to be inserted. The prompts used can be found in Fig. \ref{fig:retriever_figs}. Any text inside a set of curly brackets is a fillable part of the prompt, and any text inside angle brackets may or may not be included in a given prompt. Tags, authorities, and dates are randomly generated to fill the prompts as needed. The questions generated in this synthetic query generation procedure are split into a training group and a test group. 

\subsubsection{Evaluation Question Labeling}
    To create a set of queries labeled with relevant chunks to use to evaluate our fine-tuned retrievers, we use a similar manual labeling procedure as discussed in \ref{app:example-labeling}, but instead of having ColBERTv2 retrieve the top 20 chunks per query, it retrieves the top-50. We create a set of 50 labeled test questions in this process. As with the training query labeling, we are also able to discard irrelevant or poorly formed questions. Each of these 50 chunks per query is manually marked as relevant or irrelevant. 
    Furthermore, as we are basing the relevant chunk set on base ColBERTv2, we note that ColBERTv2 will have an inherent advantage over the fine-tuned retrievers in our evaluation process.
    \subsubsection{Retriever Evaluation}
    As discussed in \ref{app:example-labeling}, we fine-tune ColBERTv2 in 3 different ways using 3 different methods to find negative examples for each query. We evaluate these retriever variants within our RAG system using Mean Reciprocal Rank (MRR), Recall@k, and mean average precision at k (MAP@k) (for k = 5,10,20) on the set of $50$ synthetically generated and manually labeled test questions.

\subsection{System Error Analysis}
\label{app:sys_error}
We analyze system error analysis on the evaluation set in several cases.
\subsubsection{Documents Not in AGORA}
In this case, the question is "Which FY 2026 NDAA section creates (or allows creation of) AI research institutes?" The FY 2026 NDAA document is not in the AGORA dataset currently, so it should not be retrievable. However, our retriever still retrieves the NDAA from previous years, so our system mentions the establishment of AI research institutes in previous years. In some cases, the generator erroneously classifies the results from previous years as 2026, which the DPO fine-tuning helps with, as this error only occurs with the base generators.
\subsubsection{Incorrect Country Retrieval/Generation}
\label{app:err_country}
In a few cases, the question is related to one country, yet the retriever extracted context from another country, and the generator incorrectly attributed this information to the original country. For example, for the query "What role is the public sector expected to play in accelerating AI adoption in South Korea (e.g., procurement or deployment incentives)?", the retriever pulls documents from countries such as the United States, Singapore, China, and Australia. It should be noted that documents regarding the public sector in South Korea were absent in the AGORA dataset used for analysis. Though the generator is given irrelevant documents, it should still have identified that this information is not pertinent, but it instead takes the context it is given from the unrelated countries, and uses that to generate the answer to the question as if it were discussing South Korea. In other similar cases, the generator mentions that it is not given context relevant to the situation, and then discusses related documents that are retrieved in a contextually-appropriate way, making sure to mention the true source country.  
\subsection{Retriever Error Analysis}
\label{app:retrv_error}
We analyze an erroneous retrieval from each of the 3 trained retrievers.
\subsubsection{Mined negatives retriever}
The question here is "What key themes or areas of focus have emerged in AI-related policies enacted by the Commonwealth of Virginia since January 1, 2023?"
The sixth chunk retrieved by the mined negatives retriever is segment\_1459\_2. This is a segment from the West Virginia HB 5690 (Artificial Intelligence Task Force). This was proposed in 2024, but obviously is from West Virginia, not Virginia. This is an understandable mistake due to the similarity of the state names. This type of error with misidentifying authorities is a common error that the fine-tuning helped with, but is a serious concern likely to be pertinent in policy contexts.
\subsubsection{Labeled negatives retriever}
The question here is "How have pilot programs and testbeds been utilized by organizations *other than* major governmental bodies to explore and shape AI policy over time?" The first 3 retrieved chunks all come from document\_307, which is the 2023 US "Executive Order on the Safe, Secure, and Trustworthy Development and Use of Artificial Intelligence (EO 14110)". This is not relevant as it obviously is from a major governmental body. Once again, this is an issue of authority; in this case, the difficulty likely comes from the wording of the question. The authority needs to be something "*other than* major governmental bodies," which excludes rather than explicitly naming a particular authority. This requires a more nuanced interpretation of the query that is more difficult to fine-tune into the retriever.
\subsubsection{Mixed negatives retriever}
The question here is "Considering the provisions outlined in both the `Duplicative Grant Consolidation Act' and Arizona’s Senate Bill 1359, how might the use of artificial intelligence be leveraged both to prevent financial misuse *and* to address potential misinformation campaigns, particularly concerning public figures?" The first retrieved chunk is segment\_1535\_4. This is from "Utah Senate Bill 131 Information Technology Act Amendments (2024)." This obviously is not one of the documents listed in the question, and it is not entirely clear why it was the top chunk retrieved. It is possible that the query listing multiple documents and question length made accurate retrieval difficult, or that irrelevant text similarity confounded the retriever.

\section{Expert Analysis of System}
\label{sec:expert_analysis}
We provide example questions and generated answers from our system to policy researchers who are experts on AI policy. One researcher with expertise specifically in Turkey and the EU provided us with reviews of the answers to questions relating to those two documents. The following $3$ questions were provided: 
\begin{itemize}
    \item \textbf{Question 1:} Discuss the similarities and differences between Turkey's National AI strategy and the EU AI Act in terms of risk-tiering?
    \item \textbf{Question 2:} Compare Turkey's National AI Strategy proposal's comprehensiveness to other national-level AI Acts.
    \item \textbf{Question 3:} How does Turkey's National AI Strategy fall short of international standards? On which aspects can it be improved?
\end{itemize}

The following is the written feedback that we received from the policy expert, which was echoed by other policy experts as part of a workshop: 

\textbf{Assessment of Question 1 Answer}: ``Turkey's National AI strategy does not explicitly mention a risk-tiering approach, which the chatbot writes in the middle of its response. However, it starts the answer by saying "the EU AI Act and Turkey's National AI strategy both employ a risk-tiering approach". So I think the answer contradicts itself a little bit. Moreover, I wish the chatbot would capture the exact mechanism of the EU AI Act risk-tiering approach (EU AI Act categorizes AI systems into 4 groups based on risk: unacceptable risk, high risk, limited risk and minimal risk. While unacceptable risk AI systems are prohibited, high risk AI systems are subject to conformity assessment, etc.) I think this kind of answer should have been more accurate to describe the EU AI Act risk-tiering."

\textbf{Assessment of Question 2 Answer}: ``I overall liked this answer. The chatbot was able to capture the main points of the National AI strategy, and it even provided evidence from specific segments. It also rightfully pointed out that EU focuses on transparency, accountability, human oversight, and prohibiting certain uses of AI; which is lacking in Turkey. But I wished the answer was specifically pointing out that Turkey's National AI strategy does not mention ANY HARMS that could be caused during the development and use of AI systems. I think this was the main missing point in the answer.

The chatbot also pointed out that "specific actions and priorities may vary depending on the country's unique needs and circumstances," which I found interesting, because it seems like the chatbot does not take a normative stance in terms of which actions/priorities should be necessary in an AI regulatory framework."

\textbf{Assessment of Question 3 Answer}: ``I liked this answer the most and found it quite comprehensive. It was not only summarizing which aspects Turkey's document did well, but also recommending how to enhance it to meet international standards better after each specific point. (For instance, it mentioned how the strategy includes various actions to train AI experts and increase employment in the field. Then, it was giving suggestions to improve this, such as incorporating more emphasis on lifelong learning and continuous professional development to ensure the workforce remains up-to-date with the latest AI technologies and trends.)

I liked most of its suggestions as I could see it was drawing on the strengths of other international standards, but it was not really naming WHICH international standards. So, one way to improve this answer might be actually writing something like "The strategy can be improved by ensuring that AI systems are accessible to all, regardless of socio-economic background. Some international standards highlight the accessibility aspect and how AI systems should be made accessible to vulnerable groups - for example, the South Korean AI Act." In short, actually giving examples of other international standards might be quite helpful in answering a question like this, since the prompt is clearly asking how the Turkish strategy falls short of international standards."
\end{document}